\documentclass[accepted]{uai2021} 

\usepackage[greek,british]{babel}

\usepackage{mathtools} 
\usepackage{hyperref}

\usepackage{microtype}
\usepackage{graphicx}
\usepackage{subfigure}
\usepackage{booktabs} 

\usepackage{enumitem}
\usepackage{amssymb}
\usepackage{placeins}
\usepackage{natbib}

\usepackage{enumitem}
\usepackage{dblfloatfix}

\newcommand{\y}{\mathbf{y}}
\newcommand{\Y}{\mathbf{Y}}
\newcommand{\x}{\mathbf{x}}
\newcommand{\X}{\mathbf{X}}
\newcommand{\f}{\mathbf{f}}
\newcommand{\F}{\mathbf{F}}
\renewcommand{\u}{\mathbf{u}}
\newcommand{\U}{\mathbf{U}}
\newcommand{\z}{\mathbf{z}}
\newcommand{\Z}{\mathbf{Z}}

\newcommand{\K}{\mathbf{K}}
\newcommand{\m}{\mathbf{m}}
\renewcommand{\S}{\mathbf{S}}

\renewcommand{\L}{\mathcal{L}}
\renewcommand{\b}{\mathbf{b}}
\newcommand{\Q}{\mathcal{Q}}
\newcommand{\GP}{\mathcal{GP}}
\usepackage{cancel}
\usepackage[ruled,vlined, linesnumbered]{algorithm2e}
\newcommand{\I}{\mathbf{I}}
\newcommand{\0}{\mathbf{0}}

\newcommand{\N}[1]{\mathcal{N}\left(#1\right)}
\newcommand*\diff{\mathop{}\!\mathrm{d}}

\newcommand{\KL}[2]{\text{KL}[#1 \; \| \; #2]}

\DeclareMathAlphabet{\mathcal}{OMS}{cmsy}{m}{n}

\graphicspath{{figures/}} 

\makeatletter
\newcommand*{\addFileDependency}[1]{
  \typeout{(#1)}
  \@addtofilelist{#1}
  \IfFileExists{#1}{}{\typeout{No file #1.}}
}
\makeatother

\title{Probabilistic Selection of Inducing Points\\in Sparse Gaussian Processes}

\author[1]{\href{mailto:<anders.kirk.uhrenholt@gmail.com>?Subject=Your UAI 2021 paper}{Anders~Kirk~Uhrenholt}{}}
\author[1]{Valentin~Charvet}
\author[1]{Bj{\o}rn~Sand~Jensen}
\affil[1]{%
School of Computing Science, \protect\\
University of Glasgow, \protect\\
United Kingdom.
}

\begin{document}
\maketitle

\begin{abstract}
Sparse Gaussian processes and various extensions thereof are enabled through inducing points, that simultaneously bottleneck the predictive capacity and act as the main contributor towards model complexity.
%
However, the number of inducing points is generally not associated with uncertainty which prevents us from applying the apparatus of Bayesian reasoning for identifying an appropriate trade-off.
%
In this work we place a point process prior on the inducing points and approximate the associated posterior through stochastic variational inference. By letting the prior encourage a moderate number of inducing points, we enable the model to learn which and how many points to utilise.
%
We experimentally show that fewer inducing points are preferred by the model as the points become less informative, and further demonstrate how the method can be employed in deep Gaussian processes and latent variable modelling.
\end{abstract}

\section{Introduction}
Gaussian processes (GP) constitute an attractive modelling tool when the amount of data is limited and/or uncertainty quantification is critical, e.g. global optimisation (\cite{movckus1975bayesian, shahriari2015taking}), medical modelling (\cite{lorenzi2015efficient}), and reinforcement learning (\cite{deisenroth2011pilco}). While the original formulation presents shortcomings in scalability and expressiveness, a vast amount of enhancements have been presented over the years that enable large-scale modelling (\cite{Hensman2013}), deep architectures (\cite{damianou2013deep}), inter-domain covariance mappings (\cite{lazaro2009inter}), and various combinations thereof (\cite{blomqvist2019deep}).

\begin{figure*}
    \centering
    \includegraphics[width=0.85\textwidth]{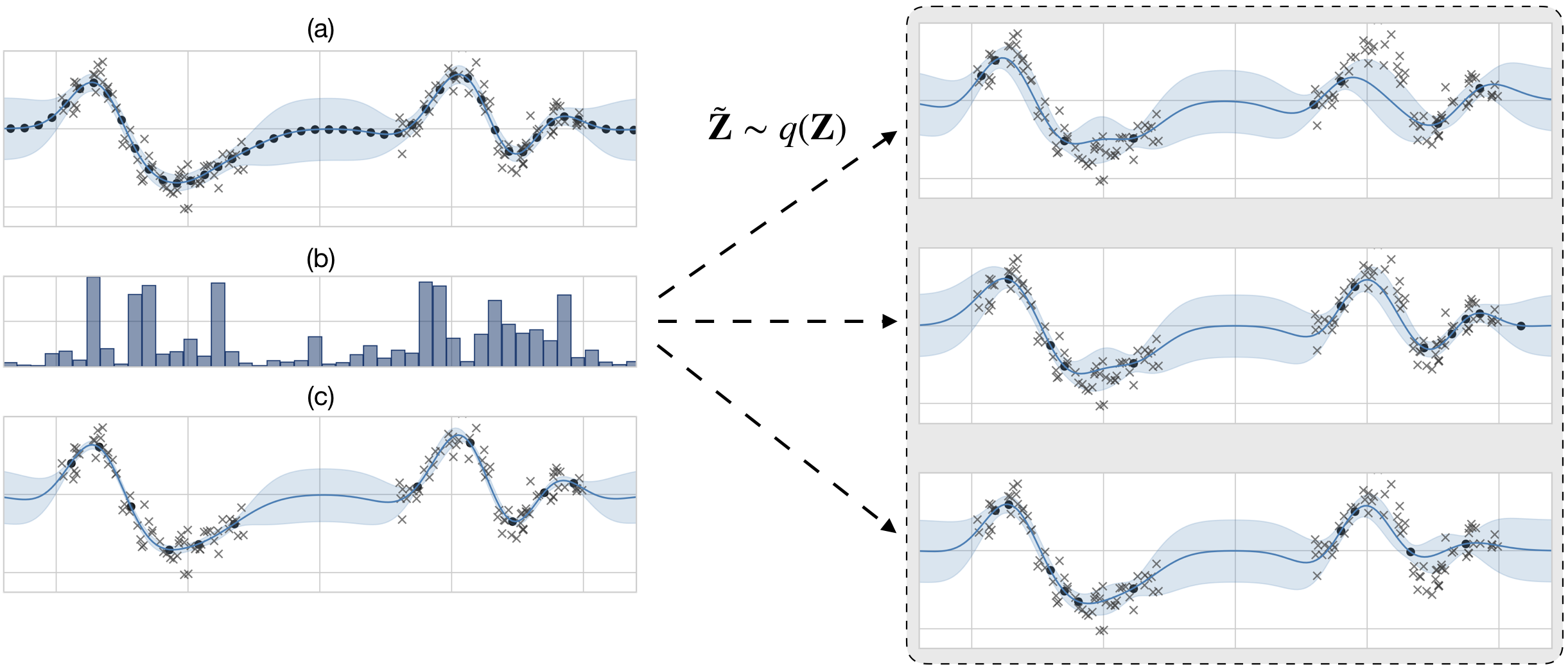}
    \vspace{-5pt}
    \caption{Illustration of our method. A sparse GP infers an approximate posterior (blue shading) over observed data (grey crosses) by conditioning on a set of inducing points (black circles). \textbf{(a)} Posterior when using an initial candidate set of 50 inducing points. \textbf{(b)} Marginal probability of inclusion assigned to each inducing point by the variational point process midway through training. Note that uninformative points are less likely to be included. Right panel shows three samples from the point process. \textbf{(c)} Result after training where a small set of highly informative inducing points remain.}
    \label{fig:simple_example}
\end{figure*}

Fundamental to these advancements is the methodology of sparse Gaussian processes (\cite{Snelson2006, Quinonero-Candela2005, pmlr-v5-titsias09a}). Here, we introduce the assumption that the true posterior over functions may be adequately approximated by conditioning on a relatively small set of \textit{inducing points}, acting as a representative proxy for the observed data (see Figure~\ref{fig:simple_example}a). The task of model optimisation now reduces to quantifying and minimising the discrepancy between the approximate posterior and that of the full GP. Since the inducing points communicate all information between observations and predictions, it is crucial that they are initialised and optimised to provide good coverage of the input domain (\cite{burt2019rates}). At the same time, they are the main contributors to the complexity of the model, and so both the number and locations of inducing points must be selected with care.

While well-developed and long underway, the inducing point methodology has always viewed the number of points as a design choice. We may add or remove inducing points adaptively and optimise their locations in input space, but the set of points ultimately appear as a deterministic quantity in the model specification. This has the immediate consequence that the question of which and how many inducing points to utilise is not part of the inference. As such, we cannot ``learn'' how many points to include in order to balance capacity and complexity, and there is no principled way of identifying inducing points that may not be contributing much towards explaining the data.

In this work we present a solution to the above limitations by including the \textit{selection} of inducing points in the inference. Concretely, we expand the hierarchical model by first sampling the inducing points from a point process prior, which in a Bayesian fashion encodes the beliefs we have about our point set -- namely that the model should be economical in its use of points. We then approximate the true posterior with a variational point process (Figure~\ref{fig:simple_example}b) that must find a trade-off between adhering to the expectations of the prior and explaining the data. As a result the model learns to sample only those points that it deems sufficiently informative, yielding a data-driven approach for optimally exploiting the available resources (Figure~\ref{fig:simple_example}c).

We present a prior that naturally encourages selectivity while allowing for an analytically tractable expression for the Kullback-Leibler divergence when certain statistical measures are known about the variational posterior. The evidence lower bound can then be maximised through score function estimation (\cite{williams1992simple, fu2006gradient}), making it straightforward to implement in any existing Gaussian process framework.\footnote{Implementation is available at \url{https://github.com/akuhren/selective_gp}.}

We demonstrate on various datasets that when reducing the informativeness of inducing points, our model does indeed select fewer points while achieving comparable or better performance than standard approaches. We then consider the problem of allocating inducing points to the layers of a deep Gaussian process, and show that we can solve this otherwise combinatorial problem by jointly inferring the posterior set of points for each layer. Finally, we apply the method to the Gaussian process latent variable model where we learn a representation for a high-dimensional single cell dataset whilst adapting the suitable number of inducing points throughout optimisation, thus avoiding costly cross-validation.

\section{Background}
\vspace{-5pt}
\label{sec:background}
\subsection{Gaussian processes}
\vspace{-5pt}
We consider the supervised learning setting of having $N$ datapoints, $(\y, \X) = \{y_i, \x_i\}_{i=1}^N$, that are assumed conditionally independent given a latent function $f$:
$$
p(\y \mid f, \X) = \prod_{i=1}^N p(y_i \mid f(\x_i)).
$$
By placing a Gaussian process prior (\cite{williams2006gaussian}) on the latent function, ${f \sim \GP\left(m(\x), \kappa(\x, \x')\right)}$, we assert that any finite set of evaluations, $\f = \{f(\x_i)\}_{i=1}^N$, follows a multivariate Gaussian with statistics given by mean function, $m$, and covariance function, $\kappa$:
\begin{gather*}
p(\f \mid \X) = \N{\f \mid \mu, \Sigma},
\\
\mu_i = m(\x_i; \theta),
\qquad
\Sigma_{ij} = \kappa(\x_i, \x_j; \theta).
\end{gather*}
Here $\theta$ comprises the hyper-parameters of $m$ and $\kappa$ that characterise the Gaussian process. When the data distribution, $p(\y \mid \f)$, is an isotropic normal, the marginal likelihood is available in closed form, providing a means for model selection and a (differentiable) objective for estimating $\theta$ and the observation noise of $p(\y \mid \f)$. Predicting for unseen data, $(\y_*, \X_*)$, is carried out by conditioning on the training data and marginalising out the latent function, yielding a tractable expression for $p(\y_* \mid \X_*, \y, \X)$.

\subsection{Sparse variational Gaussian processes}
As a result of marginalising out the latent function, the datapoints are no longer independent leading to a model that scales poorly in the number of observations. This has motivated the development of sparse Gaussian processes (\cite{Snelson2006, Quinonero-Candela2005}), in which we only condition on a set of $M$ inducing points, $(\u, \Z) = \{u_j, \z_j\}_{j=1}^M, u_j = f(\z_j)$. These points are not restricted to being part of the observed data and can thus be optimised so as to best represent $\X$ and $\y$. We will adopt the notation of $\f_{\ne \u}$ being all evaluations except for $\u$, and $\f \triangleq [\f_{\ne \u}, \u]$.
Modelling proceeds by making the simplifying assumption that the function posterior can be approximated through these inducing points, which for $M \ll N$ leads to a vast reduction in complexity. While various approaches encode this assumption in different manners, we will focus on the variational sparse formulation of \cite{pmlr-v5-titsias09a}.
By introducing a variational distribution, $\Q$, to approximate the true posterior, $p(\f \mid \y, \X)$, the evidence lower bound (ELBO) is derived through Jensen's inequality:
\begin{gather*}
\log p(\y \mid \X)
  \ge \int \Q \log \frac{p(\y, \f_{\ne \u} \mid \X)}{\Q} \diff \f_{\ne \u}
\triangleq \L.
\end{gather*}
Next, we augment the probability space with inducing points $(\u, \Z)$.\footnote{\cite{matthews2016sparse} provide a rigorous analysis of the validity of the augmentation strategy and show it to be valid under very general conditions.} Note that these points are already implicitly present in the generative model of $p(\y \mid \X)$ through the infinite-dimensional vector, $f$, but since $\f_{\ne\u}$ is a sufficient statistic for $\y$, the inducing outputs are marginalised out with no effect to the distribution over observed data. However, by including the points in the variational distribution s.t. ${\Q = p(\f_{\ne \u} \mid \X, \u, \Z)q(\u \mid \Z)}$ we obtain the much simplified expression:
\begin{align}
\L &= \int \Q \log \frac{p(\y, \f \mid \X, \Z)}{p(\f_{\ne\u} \mid \X, \u, \Z)q(\u \mid \Z)} \diff \f \nonumber
\\ &= \int \Q \log \frac{p(\y \mid \f_{\ne \u})\cancel{p(\f_{\ne \u} \mid \u, \X, \Z)}p(\u \mid \Z)}{\cancel{p(\f_{\ne \u} \mid \X, \u, \Z)}q(\u \mid \Z)} \diff \f \nonumber
\\ &= \mathbb{E}_\Q\left[\log p(\y \mid \f_{\ne \u})\right] - \text{KL}[q(\u \mid \Z) \;\|\; p(\u \mid \Z)],
\label{eq:original_ELBO}
\end{align}
where $\text{KL}[\cdot]$ is the Kullback-Leibler divergence. Our approximate posterior, $\Q \approx p(\f \mid \X, \y)$, thus relies on the initially redundant inducing points to make the function evaluations pertaining to $\y$ conditionally independent, leading to a vast reduction in complexity.

In the special case of $p(\y \mid \f_{\ne\u}) = \N{\y \mid \f_{\ne\u}, \I\sigma^2_y}$ the collapsed bound is analytically tractable and yields the variational distribution $q(\u) = \N{\u \mid \m, \S}$ with:
\begin{align*}
\S^{-1} &= \sigma^{-2}_y \K_{MM}^{-1}\K_{MN}\K_{NM}\K_{MM}^{-1} + \K_{MM}^{-1} ,
\\
\m &= \sigma^{-2}_y\S\K_{MM}^{-1}\K_{MN}\y,
\end{align*}
where $\K_{MN} = \K_{NM}^\top = \kappa(\Z, \X)$ and $\K_{MM} = \kappa(\Z, \Z)$. However, this bound does not allow us to subsample $(\y, \X)$ which in turn prohibits stochastic variational inference. To circumvent this limitation, \cite{Hensman2013} let $(\m, \S)$ be free parameters. This further enables settings where ${p(\y \mid \f_{\ne\u})}$ is not Gaussian (\cite{Hensman2015}), in which case the expectation over conditionally independent data points can be estimated by Monte Carlo sampling:
\begin{align*}
\L &= \sum_{i=1}^M \mathbb{E}_{q(f_i; \m, \S)} \left[\log p(y_i \mid f_i)\right]
\\ &\qquad - \text{KL}[q(\u \mid \Z; \m, \S) \;\|\; p(\u \mid \Z)], \nonumber
\\
q(f_i; \m, \S) &\sim \N{\beta_i\m, \kappa(\x_i, \x_i) - \beta_i(\K_{MM} - \S)\beta_i^T}, 
\\
\beta_i &= \kappa(\x_i, \Z)\K_{MM}^{-1}. \nonumber
\end{align*}

In this work we focus on both the collapsed and uncollapsed bounds and explore how the probabilistic model is affected by associating $\Z$ with a point process. The sparse variational GP (SVGP) model is straightforwardly extended to more exotic variants such as deep Gaussian processes (\cite{damianou2013deep}) and latent variable models (\cite{lawrence2004gaussian, titsias2010bayesian}), which we also explore in our experiments. However, we defer the derivation of these to Appendix~\ref{app:dgp_derivation}.

\section{Probabilistic selection of inducing points}
\label{sec:stochastic_inducing_points}
\vspace{-5pt}
In the SVGP model the inducing points communicate all information between observed data and new predictions while contributing $\mathcal{O}(M^2)$ time complexity in doing so. This implies that the main trade-off between model capacity and complexity hinges on the cardinality of $\Z$. In many scenarios this does not pose much of a dilemma. If, for instance, we are fitting a SVGP and know the amount of available training data, we would opt for using as many inducing points as possible while keeping within computational constraints. In other situations, however, the choice is less obvious. We might have a hierarchical model with various compositions of SVGP's (e.g. \cite{Hamelijnck2019}), each of which will model response surfaces of varying complexity and thus require differing amounts of inducing points. Or we may be presented with batches of data in an online fashion (\cite{bui2017streaming}) and should ideally determine adaptively how many points to utilise for a given batch dependent on the characteristics of the data.

In such situations it seems unsatisfactory to have the cardinality of $\Z$ be a deterministic choice. Rather, we ought to adopt a Bayesian approach and explicitly model any uncertainty we might have about the number of inducing points to utilise. We propose going about this by extending the generative model to include the sampling of inducing points from a point process prior, $p(\Z)$, which assigns probability to sets belonging to its domain, $\mathcal{Z}$, based on their cardinality.

This is theoretically justified since, as mentioned in the previous section, the inducing points are implicitly present in the generative model before the variational approximation is made. We are thus free to put a prior on this quantity (as has been done in previous work, e.g. \cite{hensman2015mcmc}) even though this prior does not affect the marginal distribution over $\y$. However, once the variational approximation is made the inducing points cease to be redundant and, consequently, $p(\Z)$ ceases to be redundant as well.

Note that $p(\Z)$ does \textit{not} specify a distribution over continuous values; rather it is a discrete distribution over subsets drawn from $\mathcal{Z}$. The prior will in traditional Bayesian fashion encode our expectations about any finite set of points not belonging to the observed data, allowing for a natural way to encourage a discriminatory selection of inducing points.

To accommodate this hierarchical expansion we also update our proposal distribution, $\mathcal{Q}$, to include a variational point process, $q(\Z)$:
\begin{gather*}
\mathcal{Q} = q(\f, \Z) = p(\f_{\ne \u} \mid \X, \u, \Z)q(\u \mid \Z)q(\Z).
\end{gather*}
Here, $q(\u \mid \Z)$ is just the marginal distribution conditioned on a set of inputs sampled from $\mathcal{Z}$. Under the collapsed bound this distribution is, as always, available in closed form. In the uncollapsed case, where $\mathcal{N}(\u^\star \mid \m^\star, \S^\star)$ is the variational distribution pertaining to all points in $\mathcal{Z}$ and $(\m^\star, \S^\star)$ are variational parameters, we appeal to the marginalisation properties of the normal distribution for obtaining $q(\u \mid \Z)$. That is, letting $\mathcal{I}_\Z$ be the index set associated with a particular $\Z \in \mathcal{Z}$, we have $q(\u \mid \Z) = \N{\u \mid \m^\star_{\mathcal{I}_\Z}, \S^\star_{\mathcal{I}_\Z}}$, with mean and covariance being the subvector and submatrix indexed by $\mathcal{I}_\Z$.

Including the prior and variational distribution over $\Z$ yields the updated ELBO:
\begin{align}
\log p(\y \mid \X)
  &\ge \mathbb{E}_{\mathcal{Q}}\left[
\log \frac{p(\y \mid \f_{\ne \u})p(\u \mid \Z)p(\Z)}{q(\u \mid \Z)q(\Z)}
\right]
\nonumber
\\&= \mathbb{E}_{q(\Z)}\left[\L(\Z)\right] - \text{KL}\left[q(\Z) \;\|\; p(\Z)\right]
\label{eq:updated_ELBO}
\\&\triangleq \tilde\L,
\nonumber
\end{align}
where $\L(\Z)$ is the original ELBO from \eqref{eq:original_ELBO} evaluated for a given subset $\Z \in \mathcal{Z}$. This new objective encodes the trade-off between capacity and complexity since the first term increases and the second term decreases in the number of inducing points drawn from $q(\Z)$.

\subsection{Optimising the updated ELBO}
The first term in \eqref{eq:updated_ELBO} is a discrete sum over a large (potentially infinite) domain of subsets and is therefore not straightforward to optimise. One strategy is to apply the reparameterisation trick (\cite{Maddison2016}) which allows for gradients to be propagated through samples from a continuous relaxation of the discrete distribution. For the current work we did derive an approach that utilises this method; however, as it restricts the form of $q(\Z)$ we defer it to Appendix~\ref{app:reparamerisation_trick}.

Instead we rely on the more general framework of score function estimation (\cite{williams1992simple, fu2006gradient}), which enables us to obtain unbiased, noisy gradients through Monte Carlo sampling. Letting $\lambda$ be the variational parameters of $q_\lambda(\Z)$ and noting that ${\nabla_{\lambda}q_\lambda(\Z) = q_\lambda(\Z) \nabla_{\lambda}\log q_\lambda(\Z)}$, we have:
\begin{align*}
\nabla_\lambda \mathbb{E}\left[\L(\Z)\right]
   &= \sum_{\Z\in\mathcal{Z}} \L(\Z) q_\lambda(\Z)\nabla_\lambda \log q_\lambda(\Z)
\\ &= \mathbb{E}_{q_\lambda(\Z)}[\L(\Z) \nabla_\lambda \log q_\lambda(\Z)],
\end{align*}
which can be approximated by
\begin{align}
\nabla_\lambda \mathbb{E}\left[\L(\Z)\right]
&\approx \frac{1}{S}\sum_{s=1}^S \L(\tilde\Z^s) \nabla_\lambda \log q_\lambda(\tilde\Z^{(s)}),
\label{eq:gradient_sampling}
\\
\tilde\Z^{(s)} &\sim q_\lambda(\Z). \nonumber
\end{align}
In the basic form this method may suffer from high variance in the gradients, but various strategies have been developed to alleviate this problem (\cite{glasserman2013monte}), many of which include subtracting a baseline from $\L(\tilde{\Z}^{(s)})$. We found that using a decaying average of the samples from $\L(\tilde\Z^{(s)})$ as baseline proved sufficient for obtaining stable optimisation. Note that since $\L(\Z)$ is computed as in \eqref{eq:original_ELBO}, this method can be straightforwardly implemented in any existing GP framework with automatic differentiation. The usual parameters of $\L(\Z)$ can furthermore be included in the gradient approximation of \eqref{eq:gradient_sampling} allowing for jointly learning the SVGP and its supporting inducing set.

Score function estimation increases the time complexity with a factor of $S$ compared to standard SVGP optimisation. However, when smaller sets of inducing points are sampled from $q_\lambda(\Z)$, the evaluation of $\L(\tilde\Z)$ becomes cheaper so the effective complexity depends on the degree to which the prior, $p(\Z)$, promotes sparsity.

In the following section we present point processes that make the KL divergence analytically tractable. This crucially implies that in order to maximise \eqref{eq:updated_ELBO} we only need to be able to sample from $q_\lambda(\Z)$ and evaluate the gradients of the associated log probability mass.

\subsection{Specifying the point processes}
The main motivation for this work is to allow the model to infer which inducing points to include based on their informativeness w.r.t. the observed data. We therefore find it natural to encourage fewer points by having the prior assign probability according to the squared cardinality of $\Z$:
$$
p_\alpha(\Z) = C \cdot e^{-\alpha \lvert \Z \rvert^2}.
$$
Here we have introduced $\alpha$ as a (fixed) hyper-parameter that indicates the strength of the prior while $C$ is a normalising constant. We use the squared cardinality because, for the models of our focus, the time complexity is $\mathcal{O}(M^2)$ when making new predictions. However, this is ultimately a modelling choice. Note that because we never sample from the prior when evaluating the objective in \eqref{eq:updated_ELBO}, we only have to define the probability mass function which affords us a lot of freedom in designing $p(\Z)$. Another idea that was recently explored in \cite{rossi2021sparse} is to use the prior to encourage dispersion amongst the inducing inputs. This could potentially be combined with our approach in future work.

We let $q_\lambda(\Z)$ be a discrete Poisson point process (PPP) that is constrained to a pre-defined set of candidate points, $\Z^\star$ (\cite{streit2010poisson}). This process associates each point in $\Z^\star$ with an independent probability of inclusion:
$$
q_\lambda(\Z) = \prod_{\z_k \in \Z}\lambda_k\prod_{\z_k \notin \Z}(1 - \lambda_k),
$$
with $\lambda = \{\lambda_k\}_{k=1}^K$ and 
$K$ being the total number of candidate points,
as illustrated in Figure~\ref{fig:simple_example}b. Under these definitions of $p_\alpha(\Z)$ and $q_\lambda(\Z)$, the KL divergence becomes closed-form computable. To see this, first note that $\lvert \Z \rvert$ under $q_\lambda(\Z)$ follows a Poisson binomial distribution, and so
$$
\mathcal{E} \triangleq \mathbb{E}[\lvert \Z \rvert] = \sum_{k=1}^K \lambda_k,
\;\;\;
\mathcal{V} \triangleq \text{Var}[\lvert \Z \rvert] = \sum_{k=1}^K \lambda_k(1 - \lambda_k).
$$
\begin{figure}[t!]
    \centering
    \includegraphics[width=\columnwidth]{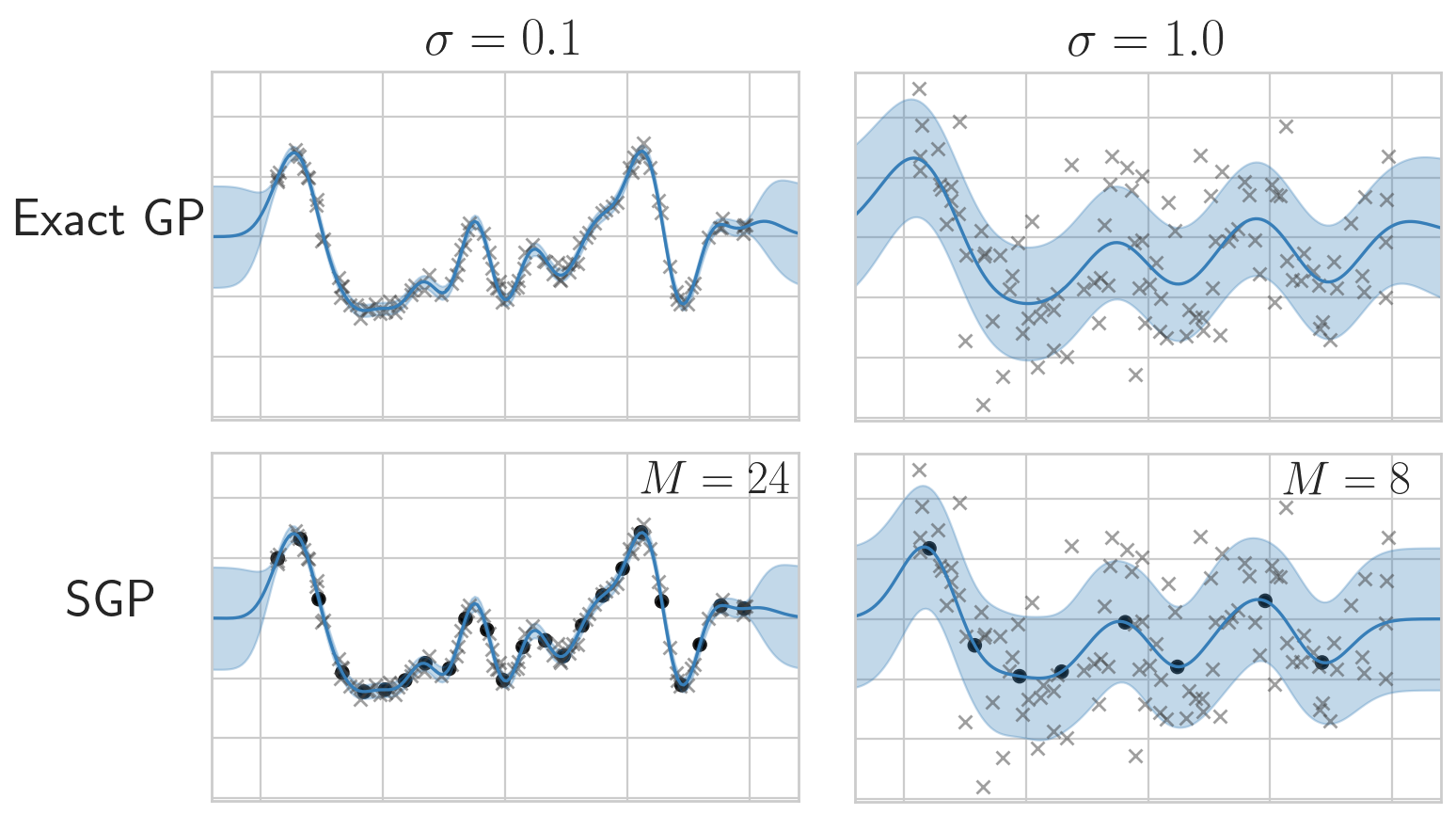}
    \caption{Interaction between observation noise, $\sigma$, and the informativeness of inducing points. As the noise increases, more of the true function is lost in the noise floor, and the exact posterior over functions (top row) becomes more uncertain. As a result, fewer inducing points are required to provide an accurate approximation of the posterior (bottom row).
    }
    \label{fig:noise_example}
\end{figure}

Decomposing into entropy and cross-entropy terms, we have:
\begin{align}
\text{KL}\left[q_\lambda(\Z) \;\|\; p_\alpha(\Z)\right] &=
 CE(\lambda, \alpha) - H(\lambda),
\label{eq:point_process_KL}
\\
H(\lambda) &= -\sum_{k=1}^K \left(\lambda_k\log \lambda_k + (1 - \lambda_k)\log \lambda_k\right),
\nonumber
\\
CE(\lambda, \alpha) &=
-\log C + \alpha\mathbb{E}_{q_\lambda(\Z)}[\lvert\Z\rvert^2]
\nonumber
\\ &= -\log C + \alpha(\mathcal{V} + \mathcal{E}^2).
\nonumber
\end{align}
The normalisation constant is ${C = 1/\sum_{k=1}^{K}\binom{K}{k} e^{-\alpha\cdot k^2}}$ but can be ignored as it contains no free parameters.

\begin{figure*}[ht!]
    \centering
    \includegraphics[width=0.97\textwidth]{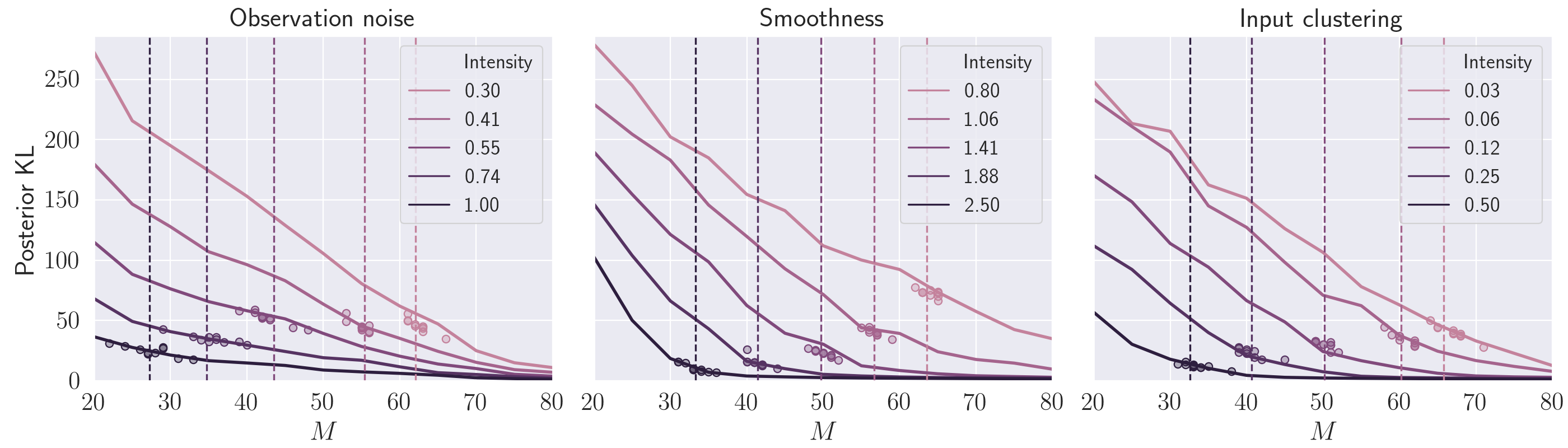}
    \caption{KL divergence between the approximate and exact posterior as a function of number of inducing points under different data characteristics (lower is better). Solid lines are the baseline SVGP with a fixed number of inducing points. As a given intensity increases, fewer points are required to obtain the same KL divergence. The vertical, dashed lines show the expected number of inducing points inferred by the point process under different characteristics, and the circles are 10 samples of subsets drawn from each. Our method adaptively reduces the number of points as informativeness decreases, and in certain cases it slightly improves upon the baseline.}
    \label{fig:noise_results}
\end{figure*}

 The new hyperparameter, $\alpha$, is user-defined and reflects the sparsity level that we expect to be sufficient for obtaining a suitable approximation. In contrast to specifying the exact number of points as in standard methods, this means that the model can identify an informative subset out of the entire set of candidates, and additionally adapt the number of points according to the amount and characteristics of the observed data.
 
\textbf{Remark:} An interesting generalisation of our framework is to let $q_\lambda(\Z)$ be a point process that is i) continuous, i.e. not restricted to a finite set of candidate points, and/or ii) capable of expressing correlation between individual points. Both of these qualities are afforded by determinantal point processes (DPP) (\cite{Kulesza2012}). However, this would render the entropy term in \eqref{eq:point_process_KL} intractable and require either that we drop it, thus loosening the bound of $\tilde{\L}$, or absorb it into the stochastic approximation of \eqref{eq:gradient_sampling}. To maintain focus on the paradigm of deriving posteriors over inducing point selection, we restrict this work to the simpler yet effective and robust  PPP setting and leave the exploration of more flexible point processes as future work.

\begin{figure*}[t]
    \centering
    \includegraphics[width=\textwidth]{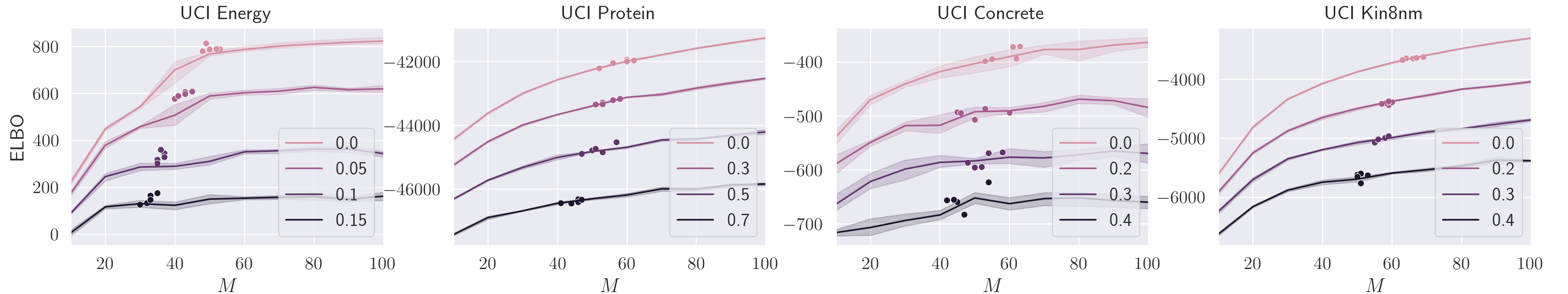}
    \caption{ELBO as a function of number of inducing points, when adding Gaussian noise of varying intensity (higher is better). The setup is similar to that of Figure~\ref{fig:noise_results}.}
    \label{fig:real_data_noise}
\end{figure*}

\section{Related work}
\label{sec:related_work}
\vspace{-10pt}
Before the introduction of inducing points, sparse GP approximations were obtained by conditioning on only a subset of the observed data (\cite{williams2001using, csato2002sparse, Lawrence2002}). Here we find commonalities with our work, since the task is to identify the most informative subset of (observed) datapoints. Generally, these methods employ discrete optimisation and often rely on greedy heuristics. 
With the introduction of inducing points the focus naturally moved to $M$ as the new bottleneck. Various strategies have been proposed for reducing complexity, e.g. decoupling the variational mean and covariance to enable distinct sets of inducing points (\cite{salimbeni2018orthogonally, havasi2018deep}) or using the spectral representation of kernels to express the posterior through Fourier features (\cite{lazaro2010sparse, hensman2017variational}).
In work parallel with ours, \cite{rossi2021sparse} also expand the hierarchical model to include the inducing inputs and even consider the DPP as a prior. Their motivation differs fundamentally from ours, however, in that they keep the number of points fixed and use the prior to improve model fitting and subsequent inference.
Identifying a good selection of inducing points has also been examined in theoretical work by \cite{burt2019rates}, who provide an asymptotic bound on the KL divergence as a function of the number of data- and inducing points, input distribution, and kernel smoothness. The latter results lend credence to common heuristics regarding how many number of inducing points to utilise for a given modelling task, and we will use them as a point of reference in our experiment section. However, they are less helpful when the influential factors are unknown prior to observing the data.

The above body of work offer various strategies for exploiting available resources when doing sparse approximations. However, a common thread is that the number of inducing points can generally be categorised as a design choice. Our contribution lies in associating this choice with uncertainty and thus reframing the problem of balancing capacity and complexity as being part of the Bayesian inference.

\section{Experiments}
\label{sec:experiments}
%
%
In this section we provide empirical evidence for the efficacy of our approach in various contexts of SVGP modelling. We first show for synthetic and real-world datasets that as the informativeness of inducing points decreases due to data characteristics, the model prunes away more points. Next, we turn to the practical problem of dynamically allocating inducing points amongst the layers of a deep Gaussian process. Lastly, we demonstrate how the method can be applied in latent variable modelling (GP-LVM) to jointly learn a low-dimensional representation along with a supporting set of inducing points.

\subsection{Informativeness of inducing points}
\label{sec:informativeness}
Previous work has identified various characteristics that determine how effective the inducing points are in communicating information about the observed data. In the works of \cite{burt2019rates} and \cite{hensman2014nested} it is shown that fewer points are required to achieve the same quality of approximation when i. the observation noise increases, ii. the kernel becomes more smooth, or iii. the input data is more clustered. This effect is shown for the case of observation noise in Figure~\ref{fig:noise_example} and illustrates how data characteristics, that are often not known a priori, influence the capacity/complexity trade-off.

\begin{figure*}[ht!]
    \centering
    \includegraphics[width=0.97\textwidth]{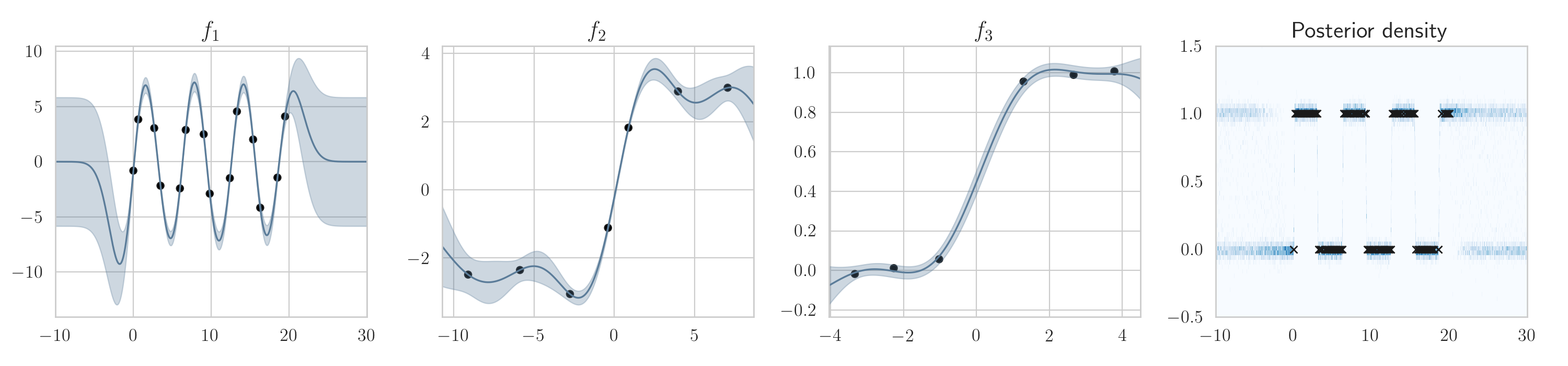}
    \vspace{-6pt}
    \caption{Illustration of different function fidelities in a 3 layer DGP. The three left plots show the learnt SVGP's for each layer with the inducing being the black circles. The target observations, drawn from a square wave, are plotted to the right on top of the DGP's posterior density.}
    \label{fig:dgp_example}
\end{figure*}

\begin{figure*}[ht!]
    \centering
    \includegraphics[width=.94\textwidth]{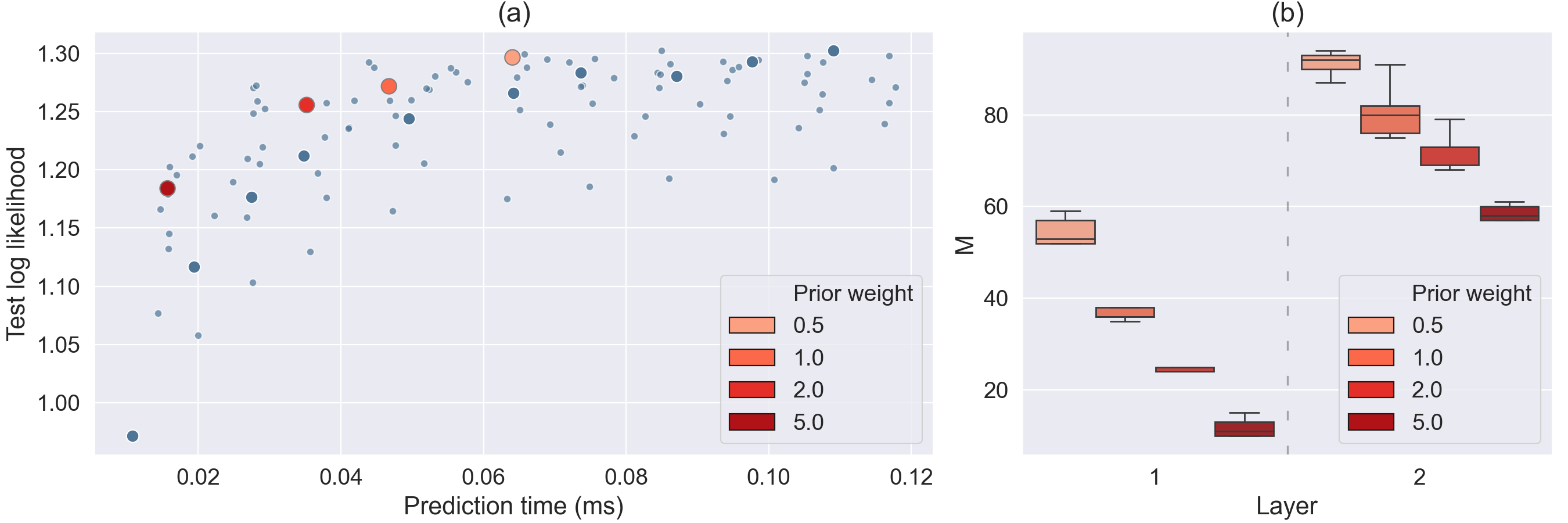}
    \vspace{-5pt}
    \caption{Dynamically allocating inducing points in a 2 layer DGP. \textbf{(a)} Each dot corresponds to the accuracy and performance of a specific configuration of inducing points. Grey dots are the outcomes of a comprehensive and tedious grid-search, with the 10 large dots representing those configurations that have equal number of inducing points both layers. Red dots are the configurations found by our point process for four different settings of the prior weight, $\alpha$. Note that the latter all fall on the frontier of optimal trade-offs between accuracy and performance. \textbf{(b)} The distribution of points across layers over 5 folds. We see a clear preference towards placing more inducing points in the last layer.
    }
    \label{fig:deep_gp_results}
\end{figure*}

In this experiment we vary the intensity of these three characteristics -- observation noise, smoothness, and input clustering -- and demonstrate that our method does indeed adapt to those conditions by selecting fewer inducing points when their informativeness decreases. As a baseline for our method we fitted SVGP's with increasing number of inducing points to noisy observations, for which we had adjusted the intensity of each characteristic separately. See Figure~\ref{fig:noise_results} where the solid lines are the posterior KL divergence as a function of number of inducing points for 5 different intensity levels. For higher intensities, the divergence reaches a plateau more quickly, showing a saturation of the information communicated by the inducing points.

We then applied our method for each intensity level, using an initial set of 80 inducing points. The vertical lines in Figure~\ref{fig:noise_results} show the expected number of points after fitting the point process for a given intensity; for each we have drawn 10 samples from the point process. The results demonstrate that the number of inferred points decrease monotonically as the intensities increase, showing that the point process does indeed adapt to the characteristics of the data; in fact, for some intensities, the adaptive approach achieves a slightly higher reduction of KL divergence.

The same experiment was repeated for 4 real-world datasets by adding varying degrees of Gaussian noise to the training outputs. Results are displayed in Figure~\ref{fig:noise_results} where we use the ELBO after 5000 epochs as metric. Again we see that as more noise is added, fewer points are deemed necessary by the points process for emulating the exact GP. The baseline for each noise level shows the trade-off between capacity and complexity that would normally have to be identified through expensive search over $M$, but that our method finds adaptively as part of the inference. We give a detailed description of the experimental setups in Appendix~\ref{app:experimental_setups}.
\begin{figure*}[ht!]
    \centering
    \includegraphics[width=1.0\textwidth]{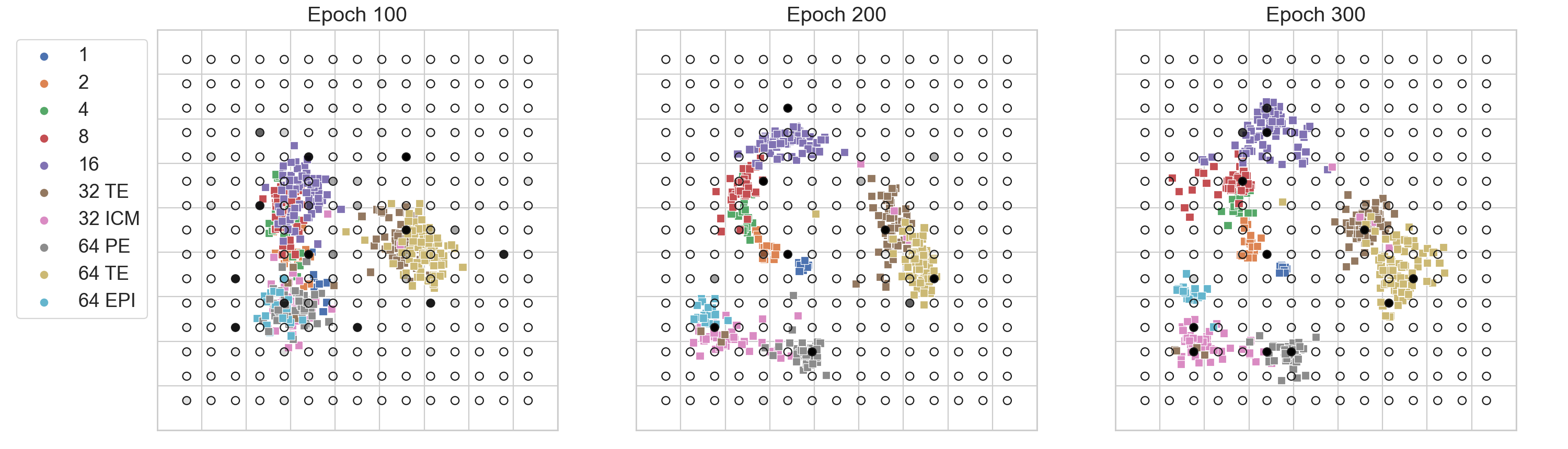}
    \vspace{-0.8cm}
    \caption{Demonstration of our method applied to a GP-LVM on a single-cell qPCR dataset where observations are coloured according to cell stage. The inducing points are fixed on a $15 \times 15$ grid with filling indicating the marginal probability of inclusion (filled means probability of one; open means probability of zero). As the latent inputs change throughout optimisation, different inducing points are activated.}
    \label{fig:gplvm}
\end{figure*}

\subsection{Deep Gaussian process}
\label{sec:dgp}
\vspace{-10pt}
When specifying a deep Gaussian process (DGP) in the sparse framework, each layer is associated with its own set of inducing points. Without evidence to the contrary, the most natural choice might be to use the same number of points for each layer. However, we put forth the conjecture that the layers could potentially model varying levels of fidelity and thus require different numbers of inducing points in order to reach an optimal trade-off between accuracy and complexity. As a motivating example, consider Figure~\ref{fig:dgp_example} where we apply a three-layer DGP to approximate a square-wave. Such discontinuous data are notoriously difficult for shallow GP's using a smooth, stationary kernel. The DGP finds a good fit by making a smooth approximation in the first layer and accentuating the jumps by step functions in the subsequent layers. Here we have applied our method to jointly learn a set of inducing points across all layers. From the figure it can be seen that the approximated functions become less complex as we move through the layers, which in turn causes fewer points to be retained by the point process.

We demonstrate that the same approach can be applied for a benchmark dataset, UCI Kin8nm (\cite{Dua:2019}), to incorporate the search for an optimum between capacity and complexity directly in the training of a two-layer DGP. The model follows the doubly-stochastic formulation from \cite{Salimbeni2017}. In accordance with the findings in \cite{duvenaud2014avoiding}, the observed input was added to the second layer which was empirically verified to increase stability. As baseline we first carried out a grid-search over $[10, 20, \dots, 90, 100]$ inducing points in each layer, yielding 100 different configurations. For each configuration we recorded the test log likelihood and prediction time over 5 folds. These results are plotted as grey dots in Figure~\ref{fig:deep_gp_results}a, and show a significant variance in prediction time for models with similar accuracy.
In our adaptive strategy we first pre-fitted the model with 150 inducing points in each layer without the point processes, estimated a posterior set of inducing points, and then re-fitted the pruned model. See Appendix~\ref{app:experimental_setups} for a detailed description. The results are plotted in Figure~\ref{fig:deep_gp_results}a for four different settings of the prior, each yielding different near-optimal trade-offs between accuracy and complexity. Figure~\ref{fig:deep_gp_results}b shows the hierarchical distribution of inducing points, illustrating a clear preference towards allocating points in the second layer. Finding such optima in the standard setting would effectively be a combinatorial problem where on would have to re-fit the model for each new configuration.

\subsection{Latent variable modelling}
\label{sec:unsupervised_setting}
\vspace{-5pt}
Lastly, we apply our method in a Bayesian Gaussian Process latent variable model (GP-LVM) where the latent inputs are learnt along with the function. We use two latent dimensions to infer a representation of 48 single cell gene expressions for 437 samples.\footnote{qPCR dataset:  \url{https://github.com/sods/ods}} The aim is to avoid the need for the extensive cross-validation required to identify a reasonable set of inducing points as the case in for example \cite{Ahmed2018}. The latent variables were initialised with PCA and the locations where fixed to a $15 \times 15$ grid for illustrative purposes. All parameters (except for inducing inputs) were trained jointly along with the point process for 300 epochs. Figure~\ref{fig:gplvm} illustrates how nearby inducing points are activated as the latent coordinates move around in the input space. In effect our method allows for the model to adapt to the change in function complexity throughout model training and negates the need for finding an appropriate number of inducing points through cross-validation.

\section{Conclusion}
\label{sec:conclusion}
\vspace{-5pt}
We have presented a Bayesian approach for choosing the number of inducing points in sparse Gaussian processes, thus introducing a probabilistic paradigm for balancing capacity and complexity. By extending the standard probabilistic model with a point process prior we encourage discriminative selection of inducing points, effectively allowing the model to choose only those points that it finds most informative. We apply variational inference to learn an approximate posterior jointly along with the usual model parameters, and present choices of prior and variational point processes that make the resulting evidence lower bound straightforward to optimise. The efficacy of this approach is verified in a controlled experiment on synthetic and real-world data, and its practical applicability is demonstrated for deep Gaussian process regression and latent variable modelling.

\vspace{0.5cm}
\textbf{Acknowledgements:} BSJ and VC acknowledge support from EPSRC grant EP/R018634/1: Closed-Loop Data Science for
Complex, Computationally- and Data-Intensive Analytics.

\bibliographystyle{abbrvnat}
\bibliography{ms}

\onecolumn
\appendix
{\LARGE\bf Supplementary material: Probabilistic selection of inducing points in sparse Gaussian processes}

\vspace{0.7cm}

\section{Derivation of deep Gaussian processes}
\label{app:dgp_derivation}
For completeness we provide a derivation of the Deep Gaussian process models used in section \ref{sec:experiments}. We follow \cite{damianou2013deep} and \cite{Salimbeni2017} and consider the layer-wise composition of functions where each function is endowed with a Gaussian process prior. By putting a prior and associated variational distribution on the inputs, the GP-LVM (\cite{lawrence2004gaussian, titsias2010bayesian}) that we consider in Section~\ref{sec:unsupervised_setting} emerges as a special case.

Considering a process with $L$ layers (that is, function compositions) with the joint likelihood:
\begin{align}
p(\Y, \{\F^\ell\}_{\ell=1}^L \mid \X)
&= p(\Y \mid \F^{L})p(\F^L \mid \F^{\ell - 1}) \dots p(\F^2 \mid \F^1)p(\F^1 \mid \X).
\label{eq:dgp_joint}
\end{align}
We capitalise all variable names to let it reflect that the function and observed outputs may have multiple dimensions. We assume w.l.o.g. that all functions have output dimensionality $D$. For notational convenience we define $\F_0 \triangleq \X$.

In our implementation, we let the function evaluations factorise across output dimensions, and so the conditional probabilities are given by:
\begin{align*}
p(\F^\ell \mid \F^{\ell - 1}) &= \prod_{1=d}^D \N{\f^{\ell, d} \mid \mu^{\ell, d}, \Sigma^{\ell, d}},
\qquad 0 \le \ell \le L,
\\
\mu^{\ell, d}_i = m^{\ell}(\F^{\ell - 1}_i)
&\qquad
\Sigma^{\ell, d}_{ij} = \kappa^\ell \left(
    \F^{\ell - 1}_i, \F^{\ell - 1}_j
\right),
\end{align*}
where $\f^{\ell, d}$ are all function evaluations of output dimension $d$, and $\F^\ell_i$ is the $i$'th evaluation across dimensions. Note that we could increase flexibility by using different kernel and mean functions for each dimension in a given layer, but for our experiments we have not found that necessary.

The posterior over latent variables in \eqref{eq:dgp_joint} is $p(\{\F\}_{\ell = 1}^L \mid \X, \Y)$ which we approximate with a variational posterior. As for the standard sparse GP, we first augment all vectors of function evaluations with inducing points, $\{\U^\ell, \Z^\ell\}_{\ell=1}^L$, which are to be marginalised out along with $\{\F^\ell\}_{\ell=1}^L$. The variational posterior, $\Q$, then takes the form:
\begin{align*}
\Q &= q(\{\F^\ell, \U^\ell\}_{\ell=1}^L)
\\ &= \prod_{\ell=1}^L p(\F^\ell \mid \U^\ell, \F^{\ell - 1}, \Z^\ell)q(\U^\ell)
\\ &= \prod_{\ell=1}^L\prod_{d=1}^D p(\f^{\ell, d} \mid \u^{\ell, d}, \F^{\ell - 1}, \Z^\ell)q(\u^{\ell, d}).
\end{align*}
Note that all dimensions for a given layer are informed by the same set of inducing inputs but different sets of inducing outputs. Inserting this into the standard ELBO derivation, we obtain:
\begin{align}
\log p(\Y \mid \X) &\ge
\int \Q \log \frac{p(\Y, \{\F^\ell, \U^\ell\}_{\ell=1}^L \mid \X, \Z)}{\Q}
\diff \{\F^\ell, \U^\ell\}_{\ell=1}^L
\nonumber
\\&=
\int \Q
\log \frac{p(\Y \mid \F^L) \{\cancel{p(\F^{\ell} \mid \U^\ell, \F^{\ell - 1}, \Z^\ell)} p(\U^\ell \mid \Z^\ell)\}_{\ell=1}^L}{\{\cancel{p(\F^{\ell} \mid \U^\ell, \F^{\ell - 1}, \Z^\ell)} q(\U^\ell)\}_{\ell=1}^L}
\diff \{\F^\ell, \U^\ell\}_{\ell=1}^L
\nonumber
\\&=
\mathbb{E}_\Q\left[\log p(\Y \mid \F^L)\right] -
    \sum_{\ell = 1}^L\sum_{d = 1}^D \KL{q(\u^{\ell, d})}{p(\u^{\ell, d} \mid \Z^{\ell})}.
\label{eq:dgp_elbo}
\end{align}
Defining $q(\u^{\ell, d}) = \N{\u^{\ell, d} \mid \m^{\ell, d}, \S^{\ell, d}}$ with $(\m^{\ell, d}, \S^{\ell, d})$ being variational parameters, the KL divergence terms of \eqref{eq:dgp_elbo} are all computable in closed form. Within each layer and for a specific dimension, the inducing outputs can be marginalised out analytically yielding a distribution that factorises over datapoints:
\begin{align*}
q(\f^{\ell, d} \mid \F^{\ell-1}, \Z^{\ell}) &=
\int p(\f^{\ell, d} \mid \u^{\ell, d}, \F^{\ell-1}, \Z^{\ell})q(\u^{\ell, d} \mid \Z^{\ell}) \diff \u^{\ell, d}
\\ &= \prod_{i=1}^N \N{f^{\ell, d}_i \mid \mu^{\ell, d}_i, \left(\sigma^{\ell, d}_i\right)^2}, 
\\
\mu^{\ell, d}_i &= m^\ell(\F_i^{\ell-1}) - \left(\alpha_i^\ell\right)^T(\m^{\ell, d} - m^\ell(\Z^\ell))
\\
\sigma^{\ell, d}_i &= \kappa^{\ell}(\F_i^{\ell-1}, \F_i^{\ell-1}) - 
\left(\alpha_i^\ell\right)^T
(\kappa^{\ell}(\Z^{\ell}, \Z^{\ell}) - \S^\ell)
\alpha_i^\ell
\\
\alpha^\ell_i &= \kappa^{\ell}(\Z^{\ell}, \Z^{\ell})^{-1}\kappa(\Z^\ell, \F_i^{\ell-1})
\end{align*}
Assuming that the likelihood is conditionally independent across both data points and dimensions, the entire expectation in \eqref{eq:dgp_elbo} decomposes into one-dimensional expectations, each of which relies only on one input observation as well as the variational parameters:
\begin{align}
\mathbb{E}_\Q\left[\log p(\Y \mid \F^L)\right]
    &= \sum_{i=1}^N\sum_{d=1}^D \mathbb{E}_{q_i^d}\left[\log p(y_i^d \mid f_i^{L, d})\right],
\nonumber
\\
q_i^d &= \prod_{\ell=1}^L \N{f^{\ell, d}_i \mid \mu^{\ell, d}_i, \left(\sigma^{\ell, d}_i\right)^2}.
\label{eq:dgp_variational_product}
\end{align}
Since each distribution in \eqref{eq:dgp_variational_product} can be re-parameterised according to \cite{kingma2015variational}, the expectation in \eqref{eq:dgp_elbo} is easily optimised through Monte Carlo sampling as suggested in \cite{Salimbeni2017}.

\vfill\pagebreak
\section{Inference using the re-parameterisation trick}
\label{app:reparamerisation_trick}
In section \ref{sec:stochastic_inducing_points} we present a general inference scheme based on score function estimation that make very few assumptions about the variational point process. However, when using a Possion point process it is possible to derive an inference scheme based on the re-parametrisaiton trick which we include here for completeness.

We build on \cite{Maddison2016} who provide a method for stochastically estimating $(\phi, \theta)$ of the expression $\mathbb{E}_{D \sim p_\phi(d)}[g_\theta(D)]$ where $D \in \{0, 1\}^L$ is a one-hot encoding of a value drawn from a multinomial distribution with probabilities $\phi = (\phi_1, \phi_2, \dots, \phi_L)$:
$$
p[D_\ell = 1] = \phi_i,\qquad\sum_{\ell=1}^L D_\ell = 1.
$$
The approach is conceptually the same as in [Kingma et al, 2015] in that the instantiations of $D \sim p_\theta(d)$ is re-written as $D = h_\phi(T), T \sim \pi(t)$ where $h_\phi$ is a discrete (and so non-differentiable), deterministic function and $\pi(t)$ is an (unparameterised) distribution over the domain of $h_\phi$. This allows us to move $\phi$ inside the expectation:
$$
\mathbb{E}_{D \sim p_\phi(d)}[g_\theta(D)] = \mathbb{E}_{T \sim \pi(t)}[g_\theta(h_\phi(T))].
$$
By further making a continuous relaxation of $h_\phi$, the expectation can be approximated and maximised through Monte Carlo sampling. This relaxation is governed by a temperature parameter, that controls how closely the continuous approximations resembles a true one-hot vector. Lower temperature means more accurate approximations but also reduces the amount of gradient information that can be communicated between states of $d$.

To apply this method for our point process estimation, we first introduce a binary vector, $\b \in [0, 1]^K$, that indicates which inducing points are present in a given sample $\Z \subseteq \Z^\star$, i.e. $b_k = 1 \Leftrightarrow \z_k \in \Z$. Next, we construct a new expression for the bound, $\hat{\L}(\Z^\star; \b)$, that takes the entire candidate set as input and masks out those inducing points for which $b_k = 0$, such that $\hat{\L}(\Z^\star; \b) = \L(\Z)$. This construction is presented in \ref{app:masking_the_bound}. If we now have a distribution over binary vectors, $q_\lambda(\b)$, that is equivalent to $q_\lambda(\Z)$ in the sense that the same subsets are given same probabilities, then:
\begin{gather}
\mathbb{E}_{q_\lambda(\Z)}\left[\L(\Z)\right]
= \mathbb{E}_{q_\lambda(\b)}\left[\hat{\L}(\Z^\star; \b)\right].  \label{eq:reparam_expectation}
\end{gather}
When $q_\lambda(\Z)$ is a discrete Poisson point process, the equivalent distribution over $\b$ is
$$
q_\lambda(\b) = \prod_{k=1}^K \lambda_k^{b_k}(1 - \lambda_k)^{1 - b_k}.
$$
Substituting this into \eqref{eq:reparam_expectation}, we have
$$
\mathbb{E}_{q_\lambda(\b)}[\hat\L(\Z^\star; \b)] =
    \mathbb{E}_{q(b_1)}[
        \mathbb{E}_{q(b_2)}[
            \cdots[
                \mathbb{E}_{q(b_K)}[\hat{\L}(\Z^\star; \b)]
            ]\cdots
        ]
    ],
$$
where $q(b_k) = \text{Bernoulli}(\lambda_k)$. Now, each of $q(b_k)$ can be re-parameterised, yielding an expression that is compatible with the formulation from \cite{Maddison2016}.

\subsection{Masking the bound}
\label{app:masking_the_bound}
The masked bound, $\hat{\L}(\Z^\star; \b)$, is obtained by evaluating the usual bound from \eqref{eq:original_ELBO} over the entire candidate set, $\Z^\star$, but under a modified GP with the following mean and kernel function:
\begin{align*}
\hat{m}_\b(\x) &=
m(\x)\cdot \Delta(\x, \b)
\\
\hat{\kappa}_\b(\x, \x') &=
\kappa(\x, \x') \cdot \Delta(\x, \b) \cdot \Delta(\x', \b) &&\text{ if $\x \ne \x'$}
\\
\hat{\kappa}_\b(\x, \x) &= \Delta(\x, \b) (\kappa(\x, \x) - 1) + 1
\\
\Delta(\x, \b) &= \prod_{j=1}^M b_j^{\delta(\x - \z_j)}.
\end{align*}
Despite the somewhat convoluted expressions, the above procedures can be carried out efficiently as vector manipulations of the mean vector and covariance matrix.

The updated mean and kernel function has the effect of factorising the complement evaluations, $\u_C = f(\Z^\star) \setminus \u$, into a standard normal:
$$
p(\y, \f, \u_C \mid \X, \Z^\star) = p(\y, \f \mid \X, \Z)p(\u_C),
\qquad p(\u_C) = \N{\u_C \mid \0, \I}.
$$
Note, that this also implies that any Gram matrix produced by $\hat{\kappa}_\b$ is positive semi-definite and so $\GP(\hat{m}_\b, \hat{\kappa}_\b)$ remains a valid Gaussian process. Assuming that a similar factorisation holds for our variational distribution s.t. $q(\f, \u_C) = q(\f)p(\u_C)$ (which is the case under e.g. the collapsed bound), we have
\begin{align*}
\L(\Z^\star; \b)
  &= \int q(\f, \u_C) \log \frac{p(\y, \f, \u_C \mid \X, \Z^\star)}
        {q(\f, \u_C)} \diff [\f, \u_C]
\\&= \int q(\f)p(\u_C) \log \frac{p(\y, \f \mid \X, \Z)p(\u_C)}
        {q(\f)p(\u_C)} \diff [\f, \u_C]
\\&= \L(\Z).
\end{align*}
As such, we have an equivalent expression for $\L(\Z)$ that lends itself to the reparameterisation trick when $q_\lambda(\b)$ is a Poisson point process.

\subsection{Comparison to score function estimation}
Empirical evidence suggests that our method, when relying on this technique for maximising \eqref{eq:updated_ELBO}, produce comparable results as the score function estimator when applied to non-deep models. However, in deep models there is a considerable propagation of noise through the layers which is induced by the continuous relaxation of the Bernoulli distributions in \eqref{eq:reparam_expectation}. This makes it difficult to get stable estimates of $\lambda$, especially in the ``deeper'' layers, prompting us to focus primarily on the score function estimator in this work.

In terms of efficiency, the re-parameterisation approach does not require multiple samples per optimisation step as is the case for SFE. However, SFE only needs to evaluate $\L(\Z)$ for those inducing points, that have been sampled from the from the candidate set, while the re-parameterisation approach always include all points. When a high level of sparsity is induced by the prior, this makes SFE notably more efficient.

\clearpage

\section{Experimental setups}
\label{app:experimental_setups}
In the following, we provide the implementation details and experimental setups necessary for recreating our results.

\subsection{Implementation}
Our implementation relies on the GPyTorch framework (\cite{gardner2018gpytorch}) which enables modular construction of exact and sparse Gaussian process models with GPU acceleration. We ran the experiments on a RTX 2080 TI GPU system. All experiments pertaining to computation speed were run on the dedicated computing server.

All of our Gaussian process priors used a zero mean function and radial basis function (RBF) kernel with Automatic Relevance Detection (ARD). We used the Adam optimiser (\cite{kingma2014adam}) with fixed learning rate of $0.01$ for all the free parameters except those related to the variational point process, which had a fixed learning rate $0.2$. In each iteration of \eqref{eq:gradient_sampling} we drew 4 samples from $q_\lambda(\Z)$.

All hyper-parameters (lengthscale and variance of the RBF kernels and, for regression, Gaussian noise variance) were initialised to 1. The inducing inputs were initialised to a random subset of observed inputs. We note, however, that we compared against other common initialisation methods (e.g. Latin hyper-cube sampling and K-means clustering) without noting any difference in performance.

All input and output data was standardised before fitting the model. However, any predictive likelihood reported was evaluated in the original rather than the scaled space.

\subsection{Training phases}
In the informativeness and DGP experiments we found it useful to divide the inference into three phases where we would first pre-train without the Poisson point process (PPP), then include the PPP in the training, sample a subset of inducing points from the PPP, and finally post-train with only those points and without the PPP. In the following we will refer to the number of epochs used for each phase as respectively $n_\text{pre}$, $n_\text{PPP}$, and $n_\text{post}$.

\subsection{Informativeness experiment}
\subsubsection{Synthetic data}
In the experiment for synthetic data of Section~\ref{sec:informativeness} we considered three generative characteristic: 1. observation noise, 2. kernel smoothness, and 3. input clustering, each of which were evaluated with different ``intensity'' levels. For each combination of condition and intensity, we sampled 500 observations from the following, generative model:
\begin{align*}
\x &\sim p(\x),
\\
\f &\sim \GP(\0, \kappa(\x, \x'; \gamma)),
\\
\y &\sim \N{\f, \I\sigma^2}.
\end{align*}
Here, $\kappa$ is an RBF kernel with variance 1.0 and lengthscale $\gamma$. To generate data for condition 1 and 2. we set each $p(\x)$ to a uniform distribution over $[0, 100]$. The intensities were then given by $\gamma$ for condition 1 and $\sigma$ for condition 2. To generate data for condition 3, we set $p(\x)$ to a homogeneous mixture of $5$ normal distributions, $\{\N{\u_j, \beta^{-1}}\}_{j=1}^5$, with equidistant means distributed over the input domain, $\mathcal{X} = [0, 100]$. The intensity was then given by the shared precision, $\beta$; i.e. higher values yields more clustering, and as $\beta \rightarrow 0$ the mixture converges weakly to a uniform distribution. The default values for the intensity parameters were $\sigma=0.1, \gamma = 1.0$.

All baselines models of the experiment were trained for 1000 epochs. For the adaptive method we fixed the prior parameter $\alpha$ to $0.05$ and used $n_\text{pre} = 200$, $n_\text{PPP} = 600$, $n_\text{post} = 200$.

\subsubsection{Real-world data}
For the real-world data we corrupted the original outputs, $\y$, with additive, standard Gaussian noise scaled by a constant factor:
$$
\hat\y = \y + \epsilon \cdot \hat\sigma(\y) \cdot v, \qquad \epsilon \sim \N{0, 1},
$$
where $\hat\sigma(\y)$ is the empirical standard deviation of the observed outputs. The constant $v \in [0, 1)$ determines the level of corruption and is the value reported in Figure~\ref{fig:real_data_noise}. The baselines models were trained for 5000 epochs. For the adaptive method we used $n_\text{pre} = 2500$, $n_\text{PPP} = 1500$, $n_\text{post} = 1000$. The prior parameters $\alpha$ needed to be configured for each dataset as more observations will tend to diminish the influence of the prior. The configurations are listed in Table~\ref{table:real_noise_spec}.

\begin{table}[h]
\centering
\begin{tabular}{|c | c | c | c | c |}
\hline
\textbf{Dataset} & \textbf{N} & \textbf{D} & \textbf{Noise levels} & $\boldsymbol\alpha$ \\ \hline
UCI Concrete & 1030 & 8 & [0.0, 0.2, 0.3, 0.4] & 0.01 \\\hline
UCI Energy & 768 & 8 & [0.0, 0.05, 0.1, 0.15] & 0.05 \\\hline
UCI Kin8nm & 8192 & 8 & [0.0, 0.2, 0.3, 0.4] & 0.1 \\\hline
UCI Protein & 45730 & 9 & [0.0, 0.3, 0.5, 0.7] & 0.2 \\\hline
\end{tabular}
\caption{Specifications for the informativeness experiment carried out on real-world benchmark datasets.}
\label{table:real_noise_spec}
\end{table}

\subsection{Deep Gaussian Process}
For all DGP experiment in Section~\ref{sec:dgp} we first trained the layers individually for 200 epochs each. In the gridsearch we then fitted both layers for another 3000 epochs. For the adaptive method we used $n_\text{pre} = 1000$, $n_\text{PPP} = 500$, $n_\text{post} = 1500$.

\subsection{GPLVM}
For the GPLVM we did \textit{not} divide into different training phases as in the other experiments. Rather, we included the PPP in the entire training and learned probability of inclusion along with the SVGP function and latent representation. The prior parameter, $\alpha$, was fixed to $3$. We used the qPCR dataset from \url{https://github.com/sods/ods} which holds 437 observations, each with 48 outputs (cell stages).

\end{document}